\documentclass[letterpaper, 10 pt, conference]{ieeeconf}  % Comment this line out if you need a4paper

\IEEEoverridecommandlockouts                              % This command is only needed if 
\title{\LARGE \bf
RoboMonitor: Label-Efficient Runtime Monitoring of Robot Task Execution via Predictive Representation Learning
}

\author{%
Abhiroop Ajith$^{1,2}$, Gokul Narayanan$^{1,\dagger}$, Kyle Coelho$^{1}$, Tingji Zhao$^{1}$, Yash Shahapurkar$^{1}$,\\
Brian Zhu$^{1}$, Melih Erdogan$^{1}$, Ted Krubasik$^{1}$, Constantinos Chamzas$^{2}$, and Eugen Solowjow$^{1}$%
\thanks{$^{1}$The authors are with Siemens Corporation, USA.
This work was conducted while Abhiroop was an intern at Siemens Corporation.}%
\thanks{$^{2}$Abhiroop and Constantinos are with the ELPIS Lab,
Worcester Polytechnic Institute, USA.}%
\thanks{\raggedright $^{\dagger}$Corresponding author: 
\texttt{gokul.sathya\_narayanan@siemens.com}}%
}

\usepackage{cite}
\usepackage{amsmath}
\usepackage{booktabs}
\usepackage{graphicx}
\usepackage{hyperref}
\usepackage{booktabs}

\begin{document}

\maketitle
\thispagestyle{empty}
\pagestyle{empty}

% --- RoboMonitor high lelvelfigure on first page (spans both columns) ---
\begin{figure*}[t]
  \centering
  \includegraphics[width=\textwidth]{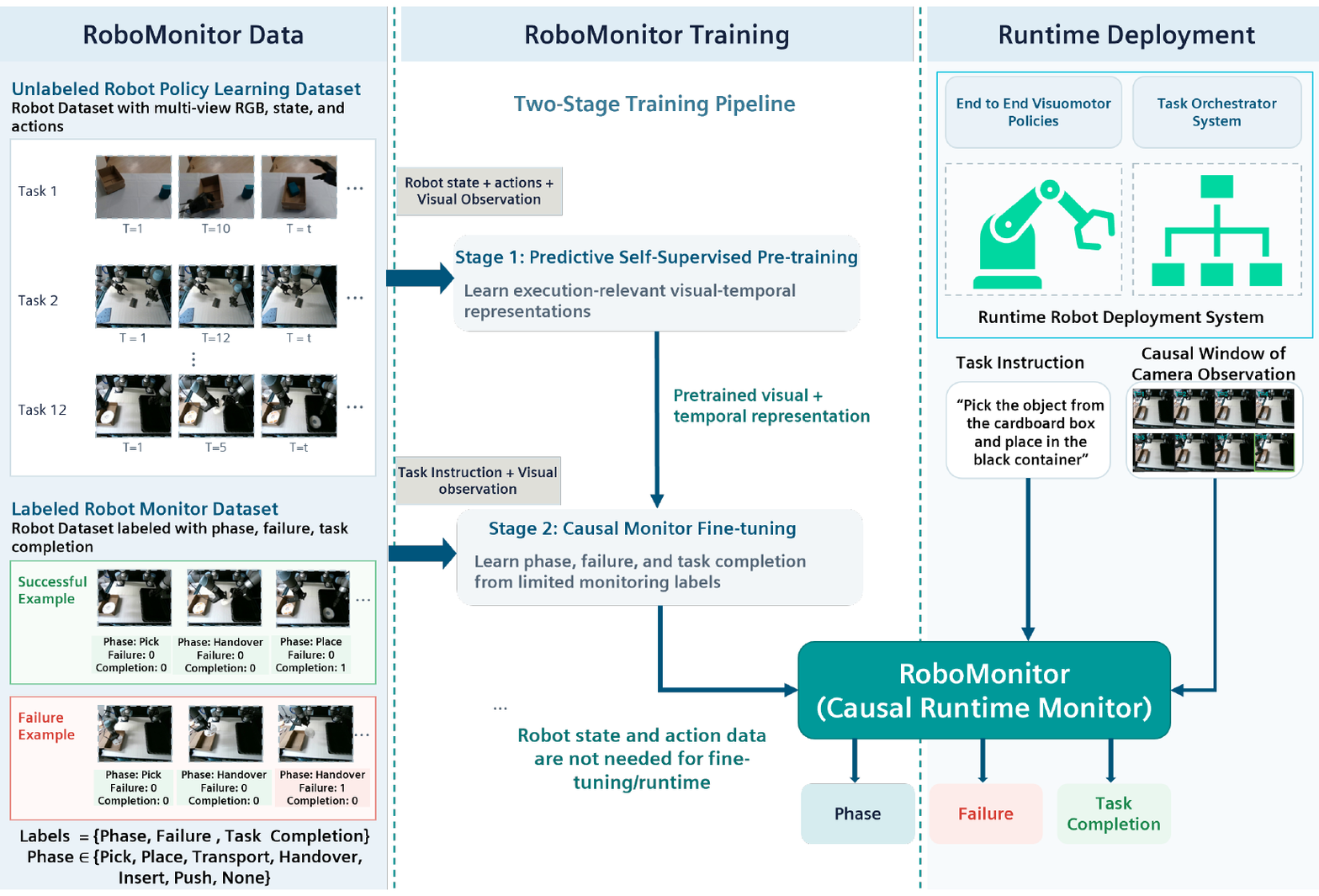}
  \vspace{-1.0em} % uncomment to tighten spacing
    \caption{\textbf{Overview of RoboMonitor.} Training has two stages: (1) predictive
    self-supervised pre-training on unlabeled robot trajectories, which uses
    multi-view observations together with recorded robot state and executed
    actions to learn execution-relevant visual and temporal representations; and
    (2) causal supervised fine-tuning (Temporal SFT) on a smaller
    monitoring-labeled dataset to predict execution phase, failure, and task
    completion. At deployment, RoboMonitor receives only the task instruction and
    camera observations, so it can monitor both end-to-end visuomotor policies and
    task-orchestration systems without access to the underlying controller.}
  \label{fig:robomonitor_overview}
\end{figure*}

\begin{abstract}
Learned robot policies produce actions, but their outputs alone do not establish whether execution is progressing as intended. Robot execution monitoring requires identifying the current execution
phase, detecting failures, and recognizing task completion from
observations available during execution. Training such monitors requires
annotations that are scarce in datasets collected for robot-policy
learning. We present RoboMonitor, a label-efficient vision--language
execution monitor that learns from these datasets before introducing
monitoring supervision. We pre-train on 25 hours of multi-camera
trajectories spanning 12 manipulation tasks and two robot embodiments,
using action-conditioned future-feature prediction, inverse dynamics,
and masked-present prediction. We then transfer the learned visual and
context encoders to a causal monitor and apply temporal supervised
fine-tuning (Temporal SFT), which combines supervision throughout each
observation window with consistency objectives within and across
overlapping windows. At deployment, RoboMonitor requires only the task
instruction and camera observations.

On a four-task monitoring benchmark, RoboMonitor trained with 52 labeled
episodes achieves 93.1\% mean phase accuracy and 85.9\% macro recall over
two fine-tuning seeds, exceeding Qwen3-VL and Robometer trained with the
same monitoring supervision. Its phase accuracy also exceeds that of both Qwen3-VL and
Robometer trained with 100 episodes. A Qwen3-VL ablation shows that
Temporal SFT reduces mean spurious phase switching from 15.23\% to
4.95\%. In closed-loop deployment, the integrated system completes
39 of 40 simulated Toolbox Sorting trials and 35 of 40 real-world Reel
Packing trials, with no false recovery triggers observed.
\end{abstract}

\section{Introduction}
\label{sec:introduction}

% Learned robot policies are increasingly used to execute long-horizon manipulation tasks~\cite{molmoact2,pi0_5,openvla}.
% They may execute an entire task end to end or serve as individual skills within an orchestration system that combines learned policies with conventional robot skills~\cite{saycan,chen2026gapgraphaspolicymultiagentselflearning,xiao2026enpireagenticrobotpolicy} to solve the task. 
% In either setting, a learned policy's action command specifies the system’s intent, but deployment also requires a mechanism to verify that the intended behavior actually occurred.
% A grasp may miss, an object may be dropped, or the outcome of a skill may differ from the transition assumed by the planner.
% Reliable deployment therefore requires an execution monitor that estimates what the robot is currently doing and whether execution is proceeding as intended.
% We consider a robot task execution monitor as a system that identifies the current execution phase, detects failures, and recognizes task completion from a task instruction and recent camera observations.
% These predictions can support task progression, recovery, replanning, or human intervention, independently of whether the
% underlying behavior is generated by an end-to-end policy or an orchestration system.

Learned robot policies are increasingly used for long-horizon
manipulation~\cite{molmoact2,pi0_5,openvla}. They can execute complete
tasks or serve as individual skills within an orchestration system
that combines learned and conventional robot
controllers~\cite{saycan,chen2026gapgraphaspolicymultiagentselflearning,xiao2026enpireagenticrobotpolicy}.
In either setting, execution requires feedback about the physical
outcome of the robot's actions. A grasp may miss, an object may be
dropped, or a skill may terminate before achieving its intended effect.
An execution monitor provides this feedback by identifying the current
phase, detecting failures, and recognizing task completion from the task
instruction and recent observations. These predictions can inform task
progression, recovery, replanning, or human intervention. Their temporal
stability also matters: phase estimates that fluctuate while the robot
remains in the same execution stage can give an orchestrator inconsistent
information about task progress.

Training such a monitor requires dense temporal annotations that are rarely available in robot-learning datasets.
Policy datasets contain large amounts of synchronized video, robot state, and action data, but are dominated by
nominal executions and generally do not indicate the current phase, the onset of a failure, or when the overall task has completed. Collecting representative failures and annotating trajectories frame by frame is costly.
However, the large amount of robot data already collected during policy development provides an opportunity to learn relevant representations of the task before monitoring annotations are introduced.
% At the same time, the robot data already collected for policy learning provide a much larger source of supervision. 
% In this work, we reuse 25 hours of unlabeled multi-camera trajectories spanning 12 manipulation tasks and two robot embodiments.

% We therefore introduce \textbf{RoboMonitor} (\autoref{fig:robomonitor_overview}), a two-stage framework that leverages unlabeled robot learning policy data to reduce the amount of monitoring-specific annotation required. First, we adapt the visual representation of a pre-trained Vision Language Model (VLM) using an action-conditioned
% joint-embedding predictive objective.
% Given recent multi-camera observations, recorded robot state, and executed commands, the model predicts future visual
% features rather than reconstructing future images. 
% We then transfer the learned visual and temporal representations to a causal VLM monitor and fine-tune it on a smaller set of trajectories annotated with monitor specific labels such as current subtask, failure, and task completion .
% Robot state and actions are used only during self-supervised pre-training; the deployed monitor receives only the task instruction and camera observations.

We introduce \textbf{RoboMonitor}
(\autoref{fig:robomonitor_overview}), a two-stage approach to
label-efficient execution monitoring. First, predictive pre-training
adapts the visual encoder of a vision--language model (VLM) and
learns a temporal context encoder from multi-camera robot trajectories. The training
objective combines future-feature prediction conditioned on recorded
robot state and commands with inverse dynamics and masked-present
prediction. We then transfer the visual and context encoders to a causal
monitor and fine-tune it on trajectories annotated with execution phase,
failure, and task completion. Our Temporal SFT procedure supervises the monitor's prediction at every timestep of each observation window rather than only at the final one, and encourages consistent predictions within and across overlapping windows. Robot state and commands are confined to pre-training; the deployed monitor uses the task instruction and camera observations.

Our contributions are: (i) a predictive pre-training and representation
transfer approach that learns from robot trajectories without monitoring
annotations; (ii) a causal execution monitor trained with dense
supervision and temporal consistency objectives; and (iii) an evaluation
of monitoring accuracy and stability at two annotation budgets, together
with closed-loop execution and recovery in simulation and on a real
robot.

\begin{figure*}[t]
    \centering
    \includegraphics[width=\textwidth]{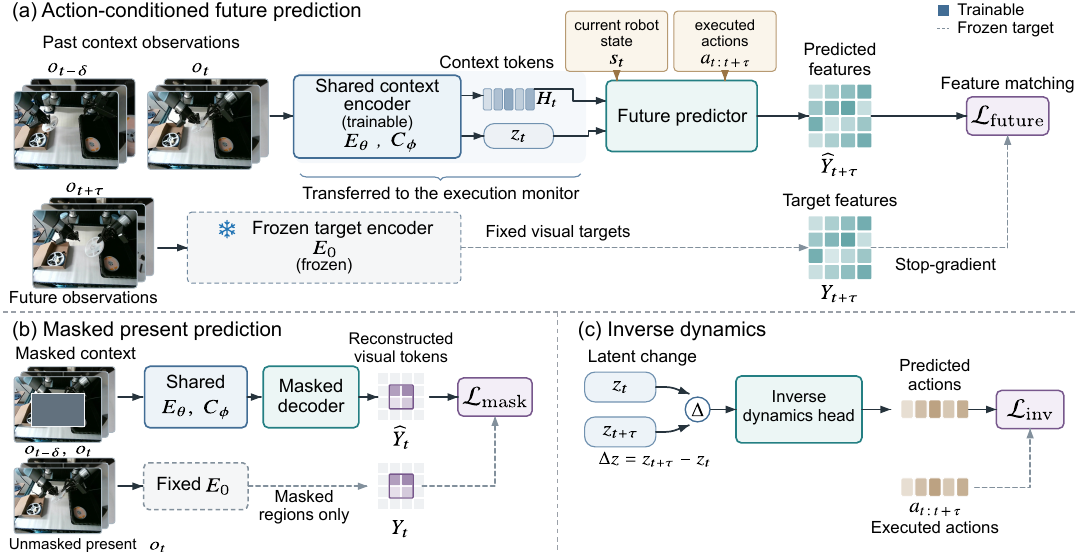}
\caption{\textbf{Predictive robot pre-training (stage 1).}
(a) Visual and context encoders process recent multi-camera observations;
a predictor uses their features, recorded robot state, and commands
to estimate future visual features. A fixed copy of the original
visual encoder supplies the targets.
(b) Masked-present prediction estimates features at masked spatial
locations.
(c) Inverse dynamics predicts commands from the difference between
context summaries. The adapted visual and context encoders are
retained for execution monitoring.}
\label{fig:ssl_pretraining}  
\end{figure*}

\section{Related Work}
\label{sec:related_work}

\noindent\textbf{Runtime Failure Detection.}
Recent work has explored failure detection using both VLM-based verification and signals from the underlying control policy. 
AHA~\cite{aha} detects and reasons about failures for a specified current subtask, while Guardian~\cite{guardian} verifies whether a given subtask was successfully executed from before-and-after observations. 
% Sentinel~\cite{sentinel} combines VLM-based progress reasoning with action-consistency signals for runtime monitoring. 
These works primarily perform subtask-conditioned verification, whereas RoboMonitor continuously infers the current execution phase, failure state, and task completion directly from causal visual observations without requiring the current subtask to be specified.
Other approaches~\cite{sentinel,fiper,safe,vlafail,armada} detect failures from probing the control policy's representations, action outputs, or comparing reference trajectories; unlike these methods, RoboMonitor is decoupled from the underlying control policy at deployment.

\noindent\textbf{Reward Models and Task Progress Estimation.}
Reward models provide dense signals about task execution.
Robometer~\cite{robometer} learns general-purpose progress and success rewards from expert, suboptimal, and failed trajectories. SARM2~\cite{sarm2} estimates the current action-primitive stage to condition a multi-task value model for policy self-improvement, while ROVER~\cite{rover} uses recursive VLM reasoning to estimate frame-level task progress over long-horizon executions. 
These methods primarily produce reward, progress, or reasoning signals for policy evaluation and improvement. In contrast, RoboMonitor treats execution monitoring as the primary objective, jointly predicting execution phase, failure, and task completion from causal visual observations, with explicit consideration of temporal stability and online inference.

\noindent\textbf{Visual and sensorimotor pre-training.}
Self-supervised pre-training has produced reusable representations for robot
learning. R3M learns visual features from human video~\cite{nair2022rm}, VIP
learns visual rewards through value-based pre-training~\cite{ma2023vip}, and
RPT jointly models robot images, proprioception, and actions before transfer
to control policies~\cite{radosavovic2023rpt}. Joint-embedding predictive
architectures learn by predicting target representations rather than
reconstructing pixels~\cite{assran2023ijepa,bardes2024vjepa}. We use this predictive principle to
learn from multi-camera robot trajectories and transfer both the adapted
visual encoder and its temporal context representation to runtime execution
monitoring.

\section{Problem Formulation}
\label{sec:problem}

We consider a manipulation task specified by an instruction $g$, executed
either by an end-to-end policy or by a high-level planner that selects
low-level skills from a library $\mathcal{K}$. At time $t$, the robot is
observed through $C$ cameras,
$o_t=(o_t^1,\ldots,o_t^C)$. Given the instruction and a causal observation
window $w_t=(o_{t-H+1},\ldots,o_t)$, the execution monitor predicts
\begin{equation}
    M_\vartheta(g,w_t)=
    \left(
        \hat{y}^{\mathrm{phase}}_t,\,
        p^{\mathrm{fail}}_t,\,
        p^{\mathrm{comp}}_t
    \right).
\end{equation}
where the outputs denote the current execution phase and the probabilities of
failure and task completion.

We assume access to a large corpus of robot trajectories containing video, state, and actions, together with a smaller corpus carrying
monitoring annotations. Our objective is to use the unlabeled
trajectories to reduce the annotations required to learn
$M_\vartheta$. At deployment, the monitor receives only the task
instruction and camera observations.

\section{Methodology}
\label{sec:methodology}
\begin{figure*}[t]
  \centering
  \includegraphics[width=\textwidth]{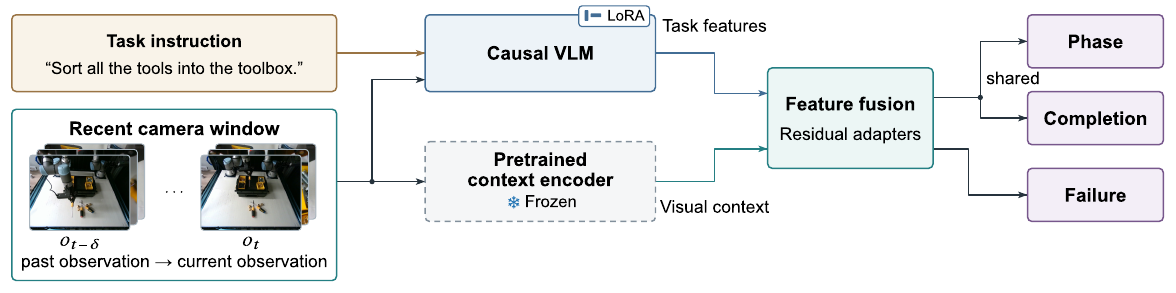}
\caption{\textbf{Causal execution monitor (stage 2).}
The VLM processes the task instruction and a multi-camera observation
window, producing a causal readout at each timestep. A separate frozen
visual/context branch supplies features from the corresponding recent
observation pair. Residual adapters combine these features using
\autoref{eq:monitor_fusion}. Phase and completion share fusion adapters;
failure uses separate adapters. Deployment uses the predictions at
the final timestep.}
\label{fig:monitor}
\end{figure*}

RoboMonitor is trained in two stages. Predictive pre-training
(\autoref{fig:ssl_pretraining}) learns visual and temporal representations
from recorded observations, robot state, and commands without monitoring
annotations. Temporal SFT (\autoref{fig:monitor}) transfers these
representations to an execution monitor and trains its phase, failure,
and completion predictions using causal observation windows.

\subsection{Predictive Robot Pre-training}
\label{sec:predictive_pretraining}

\noindent\textbf{Action-conditioned future prediction.}
Let $o_t=(o_t^1,\ldots,o_t^C)$ denote the synchronized observations
from $C$ cameras, $s_t$ the recorded robot state, and
$a_{t:t+\tau}$ the commanded action sequence over $[t,t+\tau)$,
where $\tau$ is the prediction horizon. We initialize a visual encoder $E_\theta$ from
Qwen3-VL-4B-Instruct~\cite{bai2025qwen3vl} and adapt it using visual LoRA modules~\cite{hu2022lora}. A separate copy of the original encoder,
$E_0$, provides prediction targets and remains fixed throughout pre-training. Following joint-embedding predictive learning~\cite{assran2023ijepa,bardes2024vjepa}, we predict features of
future observations rather than reconstructing their pixels.

We encode each available camera view at $t-\delta$ and $t$, where $\delta$
is a fixed context interval. The context encoder $C_\phi$ combines these
visual tokens with camera, spatial-position, and relative-time information:
\begin{equation}
    (H_t,z_t)
    =
    C_\phi\!\left(
        E_\theta(o_{t-\delta}),E_\theta(o_t)
    \right).
    \label{eq:predictive_context}
\end{equation}
Here, $H_t$ is the contextual token sequence and $z_t$ is a summary computed
from these tokens. Before future prediction, a learned projection of $z_t$
is added to $H_t$, so the future-prediction loss also trains the summary
projection.

The same observed scene can be followed by different robot commands. We
therefore condition the future predictor $P_\psi$ on the recorded state
and intervening commands:
\begin{equation}
\begin{aligned}
    \widehat{Y}_{t+\tau}
    &=
    P_\psi(H_t,z_t,s_t,a_{t:t+\tau},\tau),\\
    Y_{t+\tau}
    &=
    E_0(o_{t+\tau}).
\end{aligned}
\label{eq:future_prediction}
\end{equation}
Prediction queries specify the camera and spatial location of each target
token. Robot state and commands enter only the predictor, not $E_\theta$
or $C_\phi$. The context features can therefore be computed from camera
observations alone during monitoring.

For corresponding predicted and target tokens $\hat y$ and $y$, we use
\begin{equation}
    \ell_{\mathrm{feat}}(\hat y,y)
    = \frac{1}{d}
    \left\|\operatorname{LN}(\hat y)-\operatorname{LN}(y)\right\|_1,
    \label{eq:feature_prediction_loss}
\end{equation}
where $d$ is the token dimension and $\operatorname{LN}$ denotes layer
normalization without learned affine parameters. Averaging this loss over
valid spatial tokens, target cameras, and training examples gives
$\mathcal L_{\mathrm{future}}$.

\noindent\textbf{Inverse dynamics.}
The forward predictor receives commands as input. We additionally train an inverse-dynamics head to predict those commands from the difference between two visual context summaries~\cite{agarwal2016poking,abhiroop2026stripswm}.
Using the same encoders, we compute $z_{t+\tau}$ from observations at
$t+\tau-\delta$ and $t+\tau$, so both contexts have duration $\delta$.
The inverse prediction is
\begin{equation}
    \widehat{a}_{t:t+\tau}
    =
    I_\omega\!\left(
        z_{t+\tau}-z_t,\tau
    \right).
    \label{eq:inverse_dynamics}
\end{equation}
Unlike the forward predictor, the inverse head receives neither the recorded
state nor commands. We minimize a Huber loss over valid command entries,
$\mathcal L_{\mathrm{inv}}$, giving the summary difference an explicit
action-prediction target. The later observations used for this objective are
not inputs to the forward predictor.

\noindent\textbf{Masked-present prediction.}
The future objective predicts features at $t+\tau$. We also introduce a
prediction target at the present timestep $t$. In a separate pass, contiguous
spatial regions are masked before the visual encoder's attention layers.
An auxiliary decoder uses the remaining visual context to predict features
from $E_0$ applied to the unmasked present observation. We apply
\autoref{eq:feature_prediction_loss} only at the masked target locations,
obtaining $\mathcal L_{\mathrm{mask}}$. This decoder receives neither robot
state nor commands.

The complete pre-training objective is
\begin{equation}
    \mathcal L_{\mathrm{pre}}
    = \mathcal L_{\mathrm{future}}
    + \lambda_{\mathrm{inv}}\mathcal L_{\mathrm{inv}}
    + \lambda_{\mathrm{mask}}\mathcal L_{\mathrm{mask}},
    \label{eq:pretraining_objective}
\end{equation}
where the coefficients weight the auxiliary losses. After pre-training, we
retain $E_\theta$ and $C_\phi$ for execution monitoring. The target encoder,
future predictor, inverse-dynamics head, and masked-present decoder are not
used by the deployed monitor.

\subsection{Causal Monitor Fine-Tuning}
\label{sec:causal_monitor}
We initialize the VLM's visual encoder with the pre-trained $E_\theta$
and train the execution monitor on phase, failure, and completion
annotations. We use \emph{Temporal SFT} to denote the supervision
scheme: labeled causal readouts are trained throughout each observation
window, with additional consistency objectives within and across
overlapping windows.

\noindent\textbf{Causal readouts.}
The monitor receives the task instruction $g$ and an ordered window
$w_t=(o_{t-H+1},\ldots,o_t)$ of $H$ multi-camera observations.
After the image tokens at each timestep $j$, we insert a learned
\texttt{<|state\_reward|>} token. This \emph{readout token}
provides a dedicated position for extracting features for downstream
prediction~\cite{octo}. We call its VLM hidden state the
\emph{readout} $r_j$, which is used by the monitoring heads after
feature fusion. Causal attention restricts $r_j$ to the instruction
and observations up to timestep $j$.

Although deployment uses only the final readout, causal attention makes every
labeled position available for supervision without access to later
observations. We therefore train on all labeled readouts in each window,
rather than only the final one. The final position receives full weight,
while earlier phase and failure predictions receive a smaller context weight.
Positive failure examples receive additional weight because they are rare.

\noindent\textbf{Transferring the predictive representation.}
The VLM readout combines the task instruction with the observation
history. We augment it with features from the pre-trained context
encoder $C_\phi$. A separate frozen copy of $E_\theta$ supplies visual
features to the frozen $C_\phi$, preserving this branch while the VLM's
visual encoder is adapted through LoRA. For each supervised timestep
$j$, the context branch processes a recent observation pair ending at
$j$ and produces the corresponding tokens $H_j$ and summary $z_j$.

We form two fused features, indexed by $h\in\{\mathrm{prog},\mathrm{fail}\}$:
\begin{equation}
    \widetilde r_j^{\,h}
    = r_j + A_h(z_j) + R_h(r_j,H_j),
    \label{eq:monitor_fusion}
\end{equation}
where $A_h$ projects the summary into the VLM feature space and $R_h$
attends to the contextual tokens using $r_j$ as the query. The residual
output projections are initialized to zero, so initially
$\widetilde r_j^{\,h}=r_j$. Phase and completion share the feature indexed
by $\mathrm{prog}$; failure uses separate adapters. The output mappings are
\begin{equation}
\begin{aligned}
    \mathbf p_j^{\mathrm{phase}}
    &= \operatorname{softmax}\!\left(
        F_{\mathrm{phase}}(\widetilde r_j^{\,\mathrm{prog}})\right),\\
    p_j^{\mathrm{comp}}
    &= \operatorname{sigmoid}\!\left(
        F_{\mathrm{comp}}(\widetilde r_j^{\,\mathrm{prog}})\right),\\
    p_j^{\mathrm{fail}}
    &= \operatorname{sigmoid}\!\left(
        F_{\mathrm{fail}}(\widetilde r_j^{\,\mathrm{fail}})\right).
\end{aligned}
\label{eq:monitor_predictions}
\end{equation}
The supervised objective combines phase cross-entropy with binary
cross-entropy for failure and completion:
\begin{equation}
\begin{aligned}
    \mathcal L_{\mathrm{sup}}
    ={}& \lambda_{\mathrm{phase}}\mathcal L_{\mathrm{phase}}
       + \lambda_{\mathrm{fail}}\mathcal L_{\mathrm{fail}}\\
       &+ \lambda_{\mathrm{comp}}\mathcal L_{\mathrm{comp}}.
\end{aligned}
\label{eq:supervised_monitor_loss}
\end{equation}
Each loss incorporates the position and class weighting described above.

\noindent\textbf{Temporal consistency.}
The supervised losses compare each prediction with its annotation, but do not
directly penalize disagreement between predictions. We add two terms to
address different forms of temporal inconsistency.

Within a window, we discourage short-lived phase changes using the
truncated log-probability smoothing loss of
MS-TCN~\cite{farha2019mstcn}:
\begin{equation}
\begin{aligned}
    \Delta_{j,c}
    &= \log p^{\mathrm{phase}}_{j,c}
       - \log p^{\mathrm{phase}}_{j-1,c},\\
    \mathcal L_{\mathrm{within}}
    &= \frac{1}{|\mathcal A|K}
       \sum_{j\in\mathcal A}\sum_{c=1}^{K}
       \min\!\left(\Delta_{j,c}^{2},\kappa^{2}\right),
\end{aligned}
\label{eq:within_window_consistency}
\end{equation}
where $K$ is the number of phases, $\mathcal A$ contains positions for which
both adjacent timesteps have phase labels, and $\kappa$ is the truncation
threshold. Truncation caps the penalty on large changes rather than penalizing
them increasingly as their magnitude grows.

Overlapping windows can produce different predictions for the same
observation because they provide different preceding histories. We
penalize this disagreement at shared timestamps. For paired windows
$a$ and $b$ from the same episode, let $\mathcal O$ contain their
shared observation timestamps. Superscripts $a$ and $b$ identify
the window used to compute each prediction. We define
\begin{equation}
\begin{aligned}
    \mathcal L_{\mathrm{overlap}}
    = \frac{1}{|\mathcal O|}\sum_{u\in\mathcal O}\Big[\,
    &\tfrac12 D_{\mathrm{KL}}\!\left(
        \mathbf p_u^{a,\mathrm{phase}}\|\mathbf p_u^{b,\mathrm{phase}}\right)\\
    +{}&\tfrac12 D_{\mathrm{KL}}\!\left(
        \mathbf p_u^{b,\mathrm{phase}}\|\mathbf p_u^{a,\mathrm{phase}}\right)\\
    +{}&\beta\left(p_u^{a,\mathrm{fail}}-p_u^{b,\mathrm{fail}}\right)^2
    \,\Big],
\end{aligned}
\label{eq:cross_window_consistency}
\end{equation}
where $D_{\mathrm{KL}}$ is Kullback--Leibler divergence and $\beta$ weights
the failure term. Symmetric KL compares the phase distributions, while
squared error compares the failure probabilities.

The complete fine-tuning objective is
\begin{equation}
    \mathcal L_{\mathrm{mon}}
    = \mathcal L_{\mathrm{sup}}
    + \lambda_{\mathrm{within}}\mathcal L_{\mathrm{within}}
    + \lambda_{\mathrm{overlap}}\mathcal L_{\mathrm{overlap}}.
    \label{eq:monitor_objective}
\end{equation}
We update the monitoring heads, fusion modules, and visual and language LoRA
parameters; the remaining VLM parameters and the separate visual/context
encoders stay fixed. At deployment, the monitor returns the phase, failure,
and completion predictions for the final timestep, using only the instruction
and camera observations.

\begin{table*}[t]
\centering
\caption{Offline execution-monitoring performance.
All values are percentages; higher is better except spurious-switch
rate ($\downarrow$). Fine-tuned results are mean $\pm$ standard
deviation over two SFT seeds. Bold denotes the best mean within
each annotation budget. Pointwise SFT is evaluated only at
52 episodes. Gemini uses few-shot examples without parameter updates.}
\label{tab:robomonitor_comparison}

\scriptsize
\setlength{\tabcolsep}{2.8pt}
\renewcommand{\arraystretch}{1.15}

\begin{tabular*}{\textwidth}{
@{\extracolsep{\fill}}clccccc@{}
}
\toprule
& & \multicolumn{2}{c}{Phase} & Completion & Failure & Spurious \\\cmidrule(lr){3-4}
SFT episodes & Method
& Accuracy
& Macro Recall
& AP
& AP
& Switch Rate $\downarrow$ \\
\midrule

52
& RoboMonitor-4B (Ours)
& \textbf{93.09 $\pm$ 0.29}
& \textbf{85.85 $\pm$ 0.41}
& \textbf{87.60 $\pm$ 0.01}
& 45.87 $\pm$ 0.68
& 5.58 $\pm$ 0.22 \\

& Robometer-4B + Temporal SFT
& 88.77 $\pm$ 1.06
& 81.78 $\pm$ 0.41
& 85.42 $\pm$ 0.31
& \textbf{59.39 $\pm$ 3.11}
& 5.27 $\pm$ 0.40 \\

& Qwen3-VL-4B + Temporal SFT
& 86.76 $\pm$ 2.29
& 74.23 $\pm$ 5.69
& 74.01 $\pm$ 10.85
& 46.37 $\pm$ 6.99
& \textbf{4.95 $\pm$ 0.12} \\

& Qwen3-VL-4B + Pointwise SFT
& 85.42 $\pm$ 0.58
& 73.93 $\pm$ 0.99
& 70.14 $\pm$ 2.29
& 30.34 $\pm$ 1.16
& 15.23 $\pm$ 7.30 \\

\midrule

100
& RoboMonitor-4B (Ours)
& \textbf{93.63 $\pm$ 1.04}
& \textbf{88.55 $\pm$ 0.73}
& \textbf{92.29 $\pm$ 1.17}
& \textbf{72.01 $\pm$ 0.14}
& \textbf{3.17 $\pm$ 0.04} \\

& Robometer-4B + Temporal SFT
& 91.65 $\pm$ 0.02
& 86.22 $\pm$ 0.12
& 90.98 $\pm$ 0.01
& 67.68 $\pm$ 0.43
& 3.76 $\pm$ 0.27 \\

& Qwen3-VL-4B + Temporal SFT
& 91.19 $\pm$ 0.01
& 82.73 $\pm$ 0.64
& 84.81 $\pm$ 0.27
& 60.96 $\pm$ 0.05
& 3.58 $\pm$ 0.16 \\

\midrule

--
& Gemini Robotics ER 2 (few-shot)
& 74.67
& 61.19
& 64.07
& 32.53
& 17.62 \\

\bottomrule
\end{tabular*}
\end{table*}

\section{Experiments}
\label{sec:experiments}

We evaluate three questions:

\noindent\textbf{Q1 (Label efficiency):}
Does RoboMonitor improve execution monitoring when labeled data
are limited?

\noindent\textbf{Q2 (Temporal stability):}
Does Temporal SFT reduce spurious phase switching compared with
final-readout-only supervision?

\noindent\textbf{Q3 (Closed-loop execution):}
Can RoboMonitor support closed-loop task execution and failure recovery within a task-orchestration system?

\subsection{Data and Tasks}
\label{sec:data_tasks}

\noindent\textbf{Self-supervised corpus.}
We pre-train RoboMonitor on 25 hours of multi-camera robot
trajectories collected during policy development, spanning
12 manipulation tasks and two robot embodiments. Pre-training
uses synchronized RGB observations, recorded robot state, and
commanded actions, without phase, failure, or task-completion
annotations.

\noindent\textbf{Supervised monitoring tasks.}
We train a shared execution monitor on four manipulation tasks:

\noindent\textit{Reel Packing.}
One arm inserts its gripper into a reel to pick it up and transfers
the reel to the other arm, which places it in a black tote.
The annotated phases are \textsc{Pick}, \textsc{Handover},
and \textsc{Place}.

\noindent\textit{Electronic Component Insertion.}
The robot picks up the electronic component, transports it,
and inserts it into its testing module. The annotated phases
are \textsc{Pick}, \textsc{Transport}, and \textsc{Insert}.

\noindent\textit{Toolbox Sorting.}
The robot sorts tools and small parts, including screwdrivers,
nuts, and bolts, into designated areas in Isaac Sim. The annotated
phases are \textsc{None}, \textsc{Pick}, \textsc{Transport},
and \textsc{Place}.

\noindent\textit{Humanoid Grasping.}
The humanoid repositions a bin by pushing it to accommodate the
arms' limited reach when transferring an object between them.
The annotated phases are \textsc{Push}, \textsc{Pick},
and \textsc{Place}.

Reel Packing and Electronic Component Insertion are represented
in the self-supervised corpus, whereas Toolbox Sorting and
Humanoid Grasping are absent. All four tasks are included during
supervised fine-tuning. The monitor uses the union of their phase
labels as a shared seven-class vocabulary, with separate binary
outputs for failure and task completion.

\subsection{Offline Evaluation}
\label{sec:offline_setup}

We compare supervised training budgets of 52 and 100 labeled
episodes across the four tasks. The 100-episode set contains 27 Reel Packing, 26 Humanoid
Grasping, 25 Toolbox Sorting, and 22 Electronic Component
Insertion episodes. The 52-episode set contains
13 episodes per task and is nested within the larger set.

All methods are evaluated on the same held-out benchmark,
comprising 66 episodes and 7,022 observation windows. We
construct this benchmark to stress-test execution monitoring
across varied execution conditions. It combines nominal
executions, unsuccessful policy rollouts, and teleoperated
trajectories containing execution errors, including runs at
different controller update frequencies. This mixture tests
the models' ability to recognize execution phases and distinguish
failures from successful behavior across variation in execution
timing and rollout quality. Each evaluation query contains
the task instruction and an eight-timestep causal observation
window from the available two or three camera streams.

For each fine-tuned configuration at each evaluated budget,
we report the mean and standard deviation over two SFT seeds.
Each run is evaluated using its validation-selected checkpoint,
and all methods use the same scoring procedure.

\subsection{Baselines}
\label{sec:baselines}

We compare RoboMonitor with three fine-tuned configurations and
one few-shot embodied-reasoning baseline. Downstream SFT settings
are closely matched across the fine-tuned models, apart from the
supervision and architectural differences described below.
RoboMonitor uses a separate frozen visual/context branch and
associated fusion and readout modules, which are absent from
the Qwen3-VL and Robometer baselines.

\noindent\textbf{Qwen3-VL-4B + Pointwise SFT.}
We fine-tune Qwen3-VL-4B~\cite{bai2025qwen3vl} using the same
eight-timestep causal observation windows as the Temporal SFT
variant. Only the final readout is supervised. Intermediate
readouts receive no supervision, and the within-window and
cross-window consistency objectives are omitted. We evaluate
this ablation at the 52-episode budget.

\noindent\textbf{Qwen3-VL-4B + Temporal SFT.}
We initialize from the same Qwen3-VL-4B foundation model and
apply Temporal SFT as defined in \autoref{sec:causal_monitor}.
Comparing the Pointwise and Temporal SFT variants measures
the combined contribution of dense causal supervision and
temporal consistency.

\noindent\textbf{Robometer-4B + Temporal SFT.}
We initialize from Robometer-4B~\cite{robometer} and apply the
same Temporal SFT objective. This provides a robot-specific
reward-model baseline under the same downstream supervision.

\noindent\textbf{Gemini Robotics ER 2 (few-shot).}
We evaluate Gemini Robotics ER 2~\cite{geminiroboticser2}
without parameter updates. The model receives labeled in-context
examples together with the task instruction and causal camera
observations and predicts the same monitoring outputs.
We report this result separately from the supervised
fine-tuning budgets.

\noindent\textbf{RoboMonitor.}
RoboMonitor transfers the visual and context encoders learned
during predictive pre-training and applies Temporal SFT for
downstream execution monitoring.

\subsection{Closed-loop Evaluation}
\label{sec:online_setup}

We deploy RoboMonitor in a graph-based task orchestrator similar to GaP~\cite{chen2026gapgraphaspolicymultiagentselflearning}.
The orchestrator selects low-level skills and uses the monitor's predictions to track execution progress and trigger predefined recovery behaviors. We evaluate two settings: Toolbox Sorting with a simulated bimanual UR7e system in Isaac Sim, and Reel Packing on a real bimanual UR7e workcell. We conduct 40 trials in each setting.

\subsection{Implementation Details}
\label{sec:implementation_details}

Predictive pre-training is performed on eight NVIDIA H100 GPUs,
and supervised fine-tuning uses four GPUs. The real-world
execution monitor runs on a single NVIDIA RTX PRO 6000 Blackwell Max-Q GPU. During deployment, the monitor receives only the task instruction and camera observations; robot state and commands are not required.

\subsection{Metrics}

We report phase accuracy and macro recall, task-completion
average precision (AP), failure AP, and spurious-switch rate.
A spurious switch occurs when the predicted phase changes
between adjacent observations while the ground-truth phase
remains unchanged; lower is better. For closed-loop evaluation,
we report failure-detection recall, false recovery triggers,
recovery success, task-completion rate, and inference latency.

\section{Results}

\subsection{Label-Efficient Monitoring (Q1)}
\label{sec:label_efficiency_results}

\autoref{tab:robomonitor_comparison} reports monitoring performance
at the two annotation budgets. With 52 labeled episodes,
RoboMonitor achieves 93.09\% mean phase accuracy and 85.85\%
macro recall. These exceed Qwen3-VL with Temporal SFT by
6.33 and 11.62 percentage points, respectively, and Robometer
with Temporal SFT by 4.32 and 4.07 points. RoboMonitor also
achieves the highest task-completion AP at this budget,
87.60\%, compared with 85.42\% for Robometer and 74.01\%
for Qwen3-VL.

The cross-budget comparison further supports label efficiency
for phase recognition. RoboMonitor trained with 52 episodes
exceeds the mean phase accuracy of Robometer trained with
100 episodes (93.09\% versus 91.65\%), while its macro recall
is within 0.37 percentage points (85.85\% versus 86.22\%). RoboMonitor's 52-episode model also exceeds Qwen3-VL trained
with 100 episodes in phase accuracy (93.09\% versus 91.19\%)
and macro recall (85.85\% versus 82.73\%).
RoboMonitor therefore achieves competitive phase recognition
with fewer labeled episodes on this benchmark.

Failure detection shows a different pattern at the smaller
budget. With 52 episodes, RoboMonitor obtains 45.87\%
failure AP, below Robometer's 59.39\% and close to
Qwen3-VL's 46.37\%. With 100 episodes, RoboMonitor reaches
93.63\% phase accuracy, 88.55\% macro recall, 92.29\%
task-completion AP, and 72.01\% failure AP. All four means exceed the corresponding Robometer and Qwen3-VL
results. Failure AP is higher by 4.33 and 11.05 percentage points,
respectively. The low-label advantage
is clearest for phase recognition and completion; stronger
failure detection is observed at the larger budget.

Gemini Robotics ER 2 achieves 74.67\% phase accuracy and
61.19\% macro recall, below all fine-tuned configurations.
Its few-shot result provides a reference for monitoring with
labeled in-context examples and no parameter updates.

\subsection{Temporal Stability (Q2)}
\label{sec:temporal_stability_results}

At the 52-episode budget, Qwen3-VL with Temporal SFT reduces
mean spurious-switch rate from 15.23\% to 4.95\% relative
to Pointwise SFT, a reduction of 10.28 percentage points.
Failure AP increases from 30.34\% to 46.37\%, while phase
accuracy increases more modestly, from 85.42\% to 86.76\%.
Thus, the reduction in switching is accompanied by improved
phase accuracy and failure AP. This ablation supports the
combined dense-supervision and temporal-consistency recipe.

Across the Temporal SFT configurations, phase accuracy and
switching rate favor different methods at the smaller budget.
RoboMonitor has the highest phase accuracy with 52 episodes,
while its spurious-switch rate is 5.58\%, compared with
5.27\% for Robometer and 4.95\% for Qwen3-VL.
With 100 episodes, RoboMonitor achieves the lowest mean switching rate, 3.17\%, compared with 3.76\% for Robometer
and 3.58\% for Qwen3-VL. Switching rate should
therefore be interpreted alongside phase accuracy when
assessing execution-monitoring performance.

\begin{table}[t]
\centering
\caption{Closed-loop deployment of RoboMonitor. Recovery success
is measured over attempted recoveries.}
\label{tab:online_results}
\begin{tabular}{lcc}
\toprule
Metric & Reel Packing & Toolbox Sorting \\
       & Real & Sim \\
\midrule
Trials                  & 40              & 40 \\
Failure events          & 18              & 12 \\
Failures detected       & 15/18 (83.3\%)  & 11/12 (91.7\%) \\
False recovery triggers & 0               & 0 \\
Recovery success        & 13/15 (86.7\%)  & 11/11 (100\%) \\
Task completion         & 35/40 (87.5\%)  & 39/40 (97.5\%) \\
Inference latency       & 180--200\,ms    & 180--200\,ms \\
\bottomrule
\end{tabular}
\end{table}

\subsection{Closed-loop Execution and Recovery (Q3)}
\label{sec:closed_loop_results}

\autoref{tab:online_results} summarizes deployment within the
task orchestrator. In simulated Toolbox Sorting, RoboMonitor
detects 11 of 12 failure events across 40 trials. All
11 attempted recoveries succeed, and the integrated system
completes 39 trials (97.5\%). In real-world Reel Packing,
RoboMonitor detects 15 of 18 failure events, and 13 of
the 15 attempted recoveries succeed. Thus, 13 of the
18 encountered failures are detected and autonomously
recovered (72.2\%). The system completes 35 of 40 real-world
trials (87.5\%). No false recovery triggers are observed
in either setting.

Inference takes approximately 180--200\,ms per observation
window. These trials demonstrate the use of RoboMonitor for
task progression and predefined recovery in simulation and
on a real robot. These deployments complement the four-task offline stress test
by measuring failure-event detection, recovery, and task completion
within the integrated monitor--orchestrator system on two tasks.

\section{Conclusion}
\label{sec:conclusion}

We presented RoboMonitor, a label-efficient execution monitor that combines
predictive self-supervised pre-training with Temporal SFT. Across four
downstream manipulation tasks, RoboMonitor achieves 93.09\% mean phase
accuracy and 85.85\% macro recall with 52 labeled episodes, exceeding the
phase accuracy of the evaluated Qwen3-VL and Robometer baselines trained
with 100 episodes. An ablation of the supervised training objective further
shows that Temporal SFT reduces the Qwen3-VL spurious-switch rate from
15.23\% to 4.95\% and increases failure AP from 30.34\% to 46.37\%
relative to Pointwise SFT. In closed-loop deployment, the integrated system
completes 39 of 40 simulated Toolbox Sorting trials and 35 of 40 real-world
Reel Packing trials, with no false recovery triggers observed in either
setting.

A limitation of the current formulation is that it assigns one global
execution phase to each timestep. This represents coordinated bimanual
behaviors such as handover, but cannot express asynchronous execution in
which two manipulators independently perform different subtasks at the same
time. Future work could instead use factorized or multi-label execution
states for individual manipulators, objects, or concurrent skills. We also
plan to augment the monitor's categorical outputs with visually grounded
language reasoning~\cite{llava,ecot}, allowing the planner to receive a
description of what changed in the scene in addition to a failure score.
Finally, conformal methods~\cite{ren2023robots} could provide calibrated
uncertainty for deciding when the system should abstain, request additional
observations, or involve a human operator.
% \begin{thebibliography}{99}
\bibliographystyle{unsrt}
\bibliography{ref}

% \end{thebibliography}

\end{document}